# Automated facial recognition system using deep learning for pain assessment in adults with cerebral palsy.

Álvaro Sabater-Gárriz[a,b,c,d], F. Xavier Gaya-Morey[e] José María Buades-Rubio[c,e], Cristina Manresa-Yee[c,e] Pedro Montoya[c,d,f], Inmaculada Riquelme[b,c,d] a Balearic ASPACE Foundation, Marratxí, Spain;

b Department of Nursing and Physiotherapy, University of the Balearic Islands, Palma de Mallorca, Spain;

c Research Institute on Health Sciences (IUNICS), University of the Balearic Islands, Palma de Mallorca, Spain;

d Health Research Institute of the Balearic Islands (IdISBa) Palma de Mallorca, Spain;

e Department of Mathematics and Computer Science, University of the Balearic Islands, Palma de Mallorca, Spain;

f Center for Mathematics, Computation and Cognition, Federal University of ABC, São Bernardo do Campo, Brazil.

*Corresponding Author: inma.riquelme@uib.es

# ABSTRACT

**Background:** Pain assessment in individuals with neurological conditions, especially those with limited self-report ability and altered facial expressions, presents challenges. Existing measures, relying on direct observation by caregivers, lack sensitivity and specificity. In cerebral palsy, pain is a common comorbidity and a reliable evaluation protocol is crucial. Thus, having an automatic system that recognizes facial expressions could be of enormous help when diagnosing pain in this type of patient.

**Objectives:** 1) to build a dataset of facial pain expressions in individuals with cerebral palsy, and 2) to develop an automated facial recognition system based on deep learning for pain assessment addressed to this population.

**Methods:** Ten neural networks were trained on three pain image databases, including the UNBC-McMaster Shoulder Pain Expression Archive Database, the Multimodal Intensity Pain Dataset, and the Delaware Pain Database. Additionally, a curated dataset (CPPAIN) was created, consisting of 109 preprocessed facial pain expression images from individuals with cerebral palsy, categorized by two physiotherapists using the Facial Action Coding System observational scale.

**Results:** InceptionV3 exhibited promising performance on the CP-PAIN dataset, achieving an accuracy of 62.67% and an F1 score of 61.12%. Explainable artificial intelligence techniques revealed consistent essential features for pain identification across models.

**Conclusion:** This study demonstrates the potential of deep learning models for robust pain detection in populations with neurological conditions and communication disabilities. The creation of a larger dataset specific to cerebral palsy would further enhance model accuracy, offering a valuable tool for discerning subtle and idiosyncratic pain expressions. The insights gained could extend to other complex neurological conditions.

**Highlights**

- We construct a pioneering dataset of facial images illustrating pain in cerebral palsy.
- Our automated facial recognition system can enhance pain assessment in cerebral palsy.
- This novel system may be extrapolated to other complex neurological conditions.

# INTRODUCTION

Pain assessment is exceedingly challenging in individuals who, in addition to lacking self-report capabilities, present complex neurological conditions that impact both facial and bodily expressions. Cerebral palsy (CP) stands as a group of enduring neurological disorders impacting motor function, ranking among the most prevalent lifetime medical conditions 1,2. Its prevalence, estimated at 2-3 per 1,000 live births in developed countries 3., highlights its significant impact. Due to disruptions in brain development, individuals with CP often manifest a range of non-motor comorbidities, notably pain (74%-82%), intellectual disability (50%), speech impairment (25%), epilepsy (25%), urinary/fecal incontinence (25%), and behavioral or sleep disorders (20% to 25%) 4,5.

This chronic condition, spanning from birth throughout one's lifespan, requires ongoing therapy services 6. This demand places substantial financial burdens on families and healthcare systems and leads to extended inpatient stays 7,8. Given its chronic nature, cognitive and speech-related comorbidities, and the unique motor impairments that can affect non-verbal communication, CP serves as a valuable model for investigating complex chronic pathologies.

Pain emerges as a significant impediment to daily activities for individuals with CP 9., with a prevalence ranging from 74% to 82% 5,10. Nevertheless, diagnosing pain in this population is challenging, especially in those with cognitive or communication disorders 11. Hence, there is an urgent need for reliable and accurate pain assessment tools for individuals who cannot self-report their pain 12.

Recent years have seen various approaches to address pain assessment in non-communicative populations. These range from methods based on physiological signs to proxy observation of painful behaviors. Research has explored the utility of specific pain biomarkers such as salivary metabolites 13,14., brain activity 15,16., cardiorespiratory vital signs, skin conductance, muscle tension, or heart rate variability 17-19., for identifying the presence of acute or chronic pain.

Observational behavioral scales are the most used tools to assess pain in this population (personal communication by Sabater-Gárriz, 2023). However, their use is not without controversy, as they can yield subjective, observer-dependent data 20-22., and some may lack specificity or sensitivity 23. Observers might also confuse other emotions, such as fear or stress, for pain 20. Further, studies comparing pain assessments in children with CP by their parents have uncovered both overestimations and underestimations of pain by parents 24,25. This disparity in pain assessment extends to healthcare professionals as well 26.

In this context, the Facial Action Coding System (FACS) 27., initially designed to reduce subjectivity and provide objective descriptions of facial expressions for basic emotions, can be employed to categorize pain more objectively 28. However, mastering this system requires significant effort, and its microanalytic methods can be challenging to apply in routine clinical pain evaluations 29.

In order to automate the facial expression recognition, systems using deep learning (DL) approaches have led to significant advancements, particularly in the context of emotions, and they rely on audiovisual databases containing emotional expressions 30. Nowadays, we find multiple datasets housing collections of facial expressions depicting pain, both as standalone expressions and within broader emotional expression datasets 31,32. These datasets encompass images or videos capturing spontaneous pain, such as shoulder, neck, or lumbar pain, as well as pain induced by thermal or electrical stimuli. Some even include neonates or preschool-age children receiving injections. Within these repositories, one can find images portraying at least two levels of pain intensity, and they predominantly involve healthy individuals 32. These datasets have paved the way for the application of artificial intelligence methods, resulting in impressive achievements in pain detection, even distinguishing between spontaneous and simulated facial pain expressions 33.

Automated systems grounded in the FACS, including FaceReader 34., OpenFace 35., AFAR toolbox 36., or PainCheck® 37., have been developed for pain assessment in diverse non-communicative populations, including infants 38., and individuals with advanced dementia 39. However, these solutions may lack the specificity required to assess pain in complex neurological pathologies, such as individuals with CP, who may exhibit motor dysfunctions affecting facial expressions of pain.

Further, to understand the decision-making processes of the AI system and explore the idiosyncrasies on how it operates with individuals with CP, eXplainable Artificial Intelligence (XAI) methods can be applied 40. XAI techniques provide comprehensive explanations elucidating the decision-making processes and output generation of DL models. This facilitates the comprehension and interpretation of results by human users, enhances model trustworthiness, discerns causality among data variables or informs decision taking 40,41. In the case of facial expression recognition, we find works applying XAI techniques to emotion recognition such as sadness or happiness 42,43 or the work by Weitz et al. (2019) 44, whose research focused on the differences among facial expressions like anger or happiness, and pain.

The aim of this work is to improve pain assessment accuracy in individuals with complex neurological conditions, potentially revolutionizing the way we address their pain-related needs. In this research, our emphasis centers on deep neural networks trained using pain expression images. This focus stems from the inherent capability of these networks to adeptly handle the intricate challenges associated with recognition in real-world, or "in the wild", scenarios, as evidenced by Li and Deng (2020) 45. Our study is confined to the analysis of static images, as opposed to dynamic sequences or video data. The current study sets out to pioneer an automated facial recognition system, grounded in DL, to evaluate pain in individuals with complex neurological pathologies, specifically adults with CP. The proposed DL system will undergo training using a variety of existing pain datasets that capture diverse pain conditions. Additionally, a built purpose-designed dataset tailored for individuals with CP, denoted as CP-PAIN, will be used to evaluate the system's effectiveness.

Subsequently, the automated system's pain scores will be compared with evaluations by clinicians employing three commonly used observational scales within the CP population: The Wong Baker FACES® Pain Rating Scale 46., the Facial Action Coding System (FACS), and The Non-Communicating Adults Pain Checklist (NCAPC) 47.

Finally, to delve deeper into the mechanisms underlying DL techniques, we include eXplainable Artificial Intelligence (XAI) techniques to understand the rationale behind the outcomes of the pain perception mechanisms employed by DL models and explore potential commonalities among diverse trained DL models, especially when used with people with CP.

The work is organized as follows. Section 2 describes the methods used to build the dataset of pain expressions in individuals with CP and the automated pain recognition system. Section 3 describes the results achieved and the last Section discusses the main findings and concludes the work.

## METHODS

Within this section, we describe the methodology encompassing the construction, labelling and evaluation of a dataset of pain expressions in individuals with CP (CP-PAIN), as well as the training phases of the pain recognition artificial intelligence (AI) system.

### Constructing and Assessing the CP-PAIN Database

Prior to the construction of the CP-PAIN database, we adhered to a robust ethical framework. Written informed consent, encompassing the use of images for informative purposes, was obtained. This process involved participants with CP who had the legal capacity to provide consent (n=10), as well as legal representatives of remaining participants (n=43). The documentation was thoughtfully written in standard form and an accessible easy-to-read format to ensure its comprehension by as many participants as possible.

*Ethical Compliance and Approvals*

This research meticulously adhered to the principles outlined in the Declaration of Helsinki (1991). The research protocol received approval from both the ASPACE Foundation's ethics committee and the Research Ethics Commission of the Balearic Islands (protocol number IB4046/19), affirming its ethical rigor.

*Participants*

The study extended invitations to all users diagnosed with CP, or their legal representatives, affiliated with the Adult Life Promotion Services of the Cerebral Palsy Association (ASPACE) in the Balearic Islands (Majorca, Spain) and Toledo (Castilla-La Mancha, Spain). A total of 53 individuals (mean age=37.57 (9.88) years, age range=21-69 years, including 19 females) agreed to participate. Subsequently, they or their respective legal representatives formally filled in the informed consent.

Table 1 displays the clinical characteristics of the 53 participants with CP.

*Table 1: Clinical characteristics of participants. CP: cerebral palsy; GMCFC: Gross Motor Function Classification System 48., CFCS: Communication Function Classification System 49. These scales classify the person into 5 levels of function, lower scores indicating lower impairment of function.*

| GMFCS | n | % | CFCS | n | % | CP Subtype | n | % |
|---|---|---|---|---|---|---|---|---|
| Level I | 0 | 0 | Level I | 6 | 11.3 | Spastic | 42 | 79.2 |
| Level II | 5 | 9.4 | Level II | 7 | 13.2 | Dyskinetic | 3 | 5.7 |
| Level III | 1 | 1.9 | Level III | 11 | 20.8 | Ataxic | 1 | 1.9 |
| Level IV | 17 | 32.1 | Level IV | 14 | 26.4 | Mixed | 7 | 13.1 |
| Level V | 19 | 35.8 | Level V | 15 | 28.3 | | | |

*Measures*

In addition to gathering sociodemographic and clinical data (e.g., age, sex, type of CP, level of motor and communication impairment) from medical records, the study implemented the following measures:

*Observational Scales*

**1. The Non-Communicating Adults Pain Checklist (NCAPC)**: This 18-item scale assesses pain behaviors through six components: vocal response, emotional response, facial expression, body language, protective responses, and physiological responses. Derived from the Non-Communicating Children Pain Checklist (NCCPC), the NCAPC offers optimal utility irrespective of the evaluator's familiarity with the individual 50. The NCAPC has demonstrated strong psychometric properties and the capacity to identify pain and its intensity in adults with intellectual and developmental disabilities 47.

**2. The Wong Baker FACES® Pain Rating Scale**: This scale rates pain on a scale from 0 (no pain) to 10 (worst possible pain) by comparing the patient's facial expression to the provided scale images. While commonly used in pediatric populations, it has also found utility applied to individuals with disabilities and communication disorders 51.

**3. The Facial Action Coding System** (FACS): Designed to minimize subjective judgments in assessing facial activity, FACS is a widely employed system for coding emotional facial expressions in scientific studies 52. It has been successfully applied to individuals with communication disorders and CP 53. FACS dissects facial expressions into 44 individual components of muscle movement, known as Action Units (AUs), rating them on a 6-point scale (0= no expression, 5= extreme expression). Pain is identified through six specific AUs: lowering of the eyebrows (AU4), elevation of the cheeks and compression of the eyelids and/or contraction of the cheekbones (AU6/AU7), wrinkling of the nose and/or raising the upper lip (AU9/AU10), and closing the eyes (AU43) 29,54.

The total pain score is calculated as follows: Pain score= AU4 + (AU6||AU7) + (AU9||AU10) + AU43, yielding a 20-point scale.

*Procedure for Data Collection*

Video recordings capturing facial and body expressions were conducted in situations where participants with CP either underwent potential painful procedures or when caregivers identified signs of pain in other care procedures (e. g. feeding, personal hygiene, assistive...). For scheduled painful procedures such as therapies or intramuscular injections, the video recording initiated 2 minutes prior to the potentially painful stimuli and continued for 2 minutes thereafter. In cases where participants had the cognitive capacity (n=24, resulting in a total of 43 recorded expressions), they were kindly asked to self-rate their pain on a scale of 1 to 10 using the Wong-Baker Faces pain rating scale.

Throughout the study duration, a comprehensive total of 127 recordings were successfully acquired. These recordings depicted various sources of pain in the painful images, which were classified as follows:

- Intramuscular injection: 77 images (60.6%)

- Muscular stretching: 32 images (25.2%)

- Other sources of pain: 18 images (14.2%)

*Expert Evaluation of Video Recordings*

All video recordings underwent meticulous offline evaluation by two highly experienced physiotherapists, with more than 10 years of expertise in treating individuals with CP. These evaluators independently applied the three observation scales: the Wong-Baker Faces pain rating scale, NCAPC, and FACS.

To analyze the agreement between the two raters, the Intraclass Correlation Coefficient (ICC) was computed for each scale. Physiotherapists might be more accurate recognizing pain expressions resulting from their own interventions, such as muscle stretching, which they have observed on a regular basis, rather than pain expressions caused by procedures less related to their profession, such as intramuscular injections. Thus, the ICC was separately calculated for each type of painful stimulus, in order to ascertain whether the evaluator's familiarity with the particular painful procedure or situation influenced the agreement.

The interpretation of the ICC scores followed the categories proposed by Fleiss (1986) [55.,]: low agreement (ICC < 0.40), fair/good agreement (ICC 0.41 to 0.75), and excellent agreement (ICC > 0.75).

For a comprehensive comparison with the AI system's classification, we transformed the mean scores for each observational scale into a binary format. Specifically, a score of '0' indicated 'no pain,' while any score greater than 0 indicated the presence of pain. Therefore, to be labelled as "no pain," an image required a unanimous '0' score agreement between both evaluators. In all other cases, the image was labelled as 'pain'.

*Challenges with Self-Reports*

Despite our initial efforts to collect retrospective self-reports from participants after the conclusion of the video recordings, we encountered notable challenges. The primary obstacle stemmed from cognitive and attentional issues experienced by the participants. Consequently, only a limited number of self-reports, totaling 10, could be successfully gathered.

## Deep learning pain recognition

*Datasets*

Aiming at building a pain recognition system applicable to images of individuals with CP, we merge three extensively used databases: the UNBC-McMaster Shoulder Pain Expression Archive Database [56.], the Multimodal Intensity Pain dataset (MInt PAIN), and the Delaware Pain Database [31]. For the sake of brevity, we will subsequently denote this merged dataset as PAIN-DB.

The UNBC-McMaster Shoulder Pain Expression Archive Database comprises 25 adult patients afflicted with shoulder pain. It features a collection of 200 distinct range of motion tests, encompassing both affected and unaffected limbs. Data acquisition involves videos capturing facial expressions, albeit in low resolution, which also include social interaction and verbal communication. Annotations include self-report measurements via Visual Analog Scales (VAS) encompassing sensory and affective aspects, along with pain intensity assessed by both self-report and observer (Observer-Assessed Pain Intensity, OPI). Moreover, annotations encompass limb information (affected/unaffected) and FACS coding.

The MInt PAIN Database presents a collection of images obtained through electrical muscle pain stimulation, involving 20 subjects. Each subject participated in two trials during data acquisition, with each trial encompassing 40 pain stimulation sweeps. Within these sweeps, two types of data were captured: one representing the absence of pain, and the other portraying pain at four varying intensities.

The Delaware Pain Database is an extensive compilation of fully characterized photographs featuring 127 female and 113 male subjects, with an emphasis on painful and neutral expressions. The dataset's primary hallmark lies in its remarkable diversity across dimensions of race, gender, and expression intensity.

Upon the integration of the three datasets, a relabeling into two classes was performed: images featuring pain and images devoid of pain. Consequently, images encompassing varying degrees of pain were grouped within the first class, while the remaining images were categorized under the second class. This reduction served two primary objectives: foremost, the normalization of data across the datasets, aligned with the two-class structure in the Delaware Pain Database; and secondly, the transformation of the task into a binary classification problem.

Furthermore, we did not use all available frames from UNBC-McMaster and MInt PAIN, to avoid overfitting on the users they contain. Since there is little variation in consecutive frames, instead of using them all, a sample of twenty frames per user and class was taken. The final PAIN-DB dataset, composed by images from the three described databases, is summarized in Table 2. Figure 1 depicts some example images from each database.

*Table 2: Breakdown of PAIN-DB dataset into the datasets forming it.* Last column refers to the number of pain levels used to label the images of the dataset. Notice that not all images from the MInt PAIN and the UNBC-McMaster datasets were used, and that the final dataset contains only two classes: pain and no pain.

| Dataset | Users | Images | Levels |
| --- | --- | --- | --- |
| MInt | 20 | 800 | 5 |
| Delaware | 240 | 803 | 2 |
| UNBC-McMaster | 25 | 980 | 5 |
| Total | 285 | 2.583 | 2 |

*Figure 1: Example of images with and without pain, from the four datasets employed in this study.*

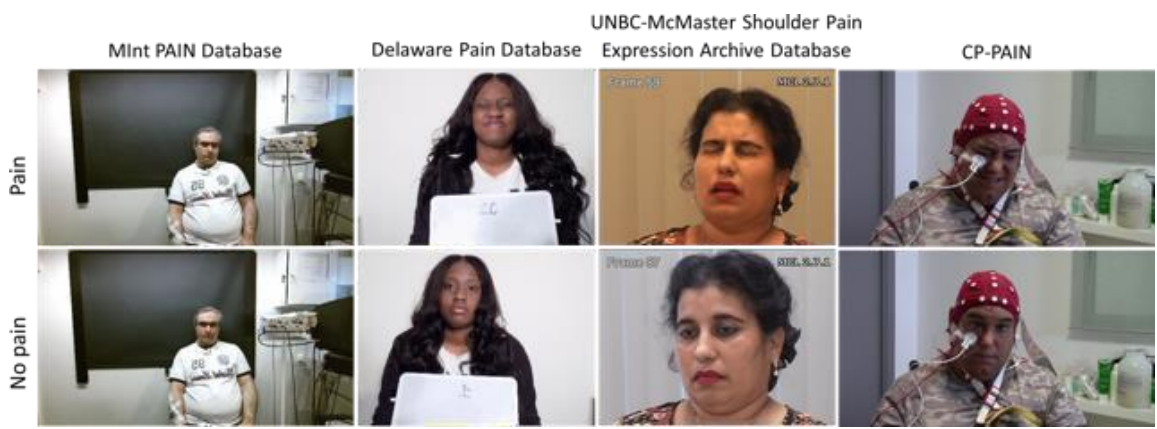

To assess the efficacy of the PAIN-DB-trained pain recognition system on individuals with CP, a testing dataset was curated from video recordings of study participants with CP, designated as CP-PAIN. By using pin-pointed moments of pain expression as determined by one physiotherapist specialized in neurological rehabilitation, two frames were extracted from each video: one capturing the moment of pain manifestation and another from a non-pain moment, thus ensuring a balanced dataset composition. Regrettably, a subset of videos proved unsuitable for inclusion due to two primary reasons: substantial occlusions obscuring facial features during pain instances, and participants wearing surgical masks that obscured half of the face. Although human observers can bypass these occlusions to discern pain expressions, our automated system lacked training for such scenarios, which could introduce unpredictable outcomes. Consequently, these videos were omitted. The resulting CP-PAIN dataset comprised 109 images, representing a dataset suitable for evaluation. Two example images from this dataset can be found on Figure 1, in the rightmost column.

*Image preprocessing*

We conducted a pre-processing phase, to enhance the performance of the neural networks, which comprised cropping and background subtraction.

Images were cropped into a squared region containing the face (similar to Haque et al. (2018) approach [57]). This was achieved through the application of the well-established multitask cascaded convolutional networks for face detection [58].

Furthermore, the background subtraction was accomplished through the integration of U2-Net [59]., which facilitates precise object segmentation. In our case, we used their pretrained weights that had been fine-tuned for human segmentation. Following the extraction of a binary mask outlining the background pixels, these regions were replaced with white color, thereby retaining only the human subject within the image frame.

The combined cropping and background subtraction procedures were devised to streamline the pain recognition task for the neural networks. This dual process serves both to eliminate any interference introduced by factors other than the person's facial features and to standardize the images. This standardization ensures that the facial attributes consistently occupy an approximate spatial alignment. Finally, the resultant images were resized to the input dimensions required by the distinct network architectures utilized in our study. Examples of the preprocessed images for each class and dataset are shown in Figure 2.

**Figure 2: Resulting images after the pre-processing performed before any training or testing on the four databases, featuring a face crop and background subtraction.**

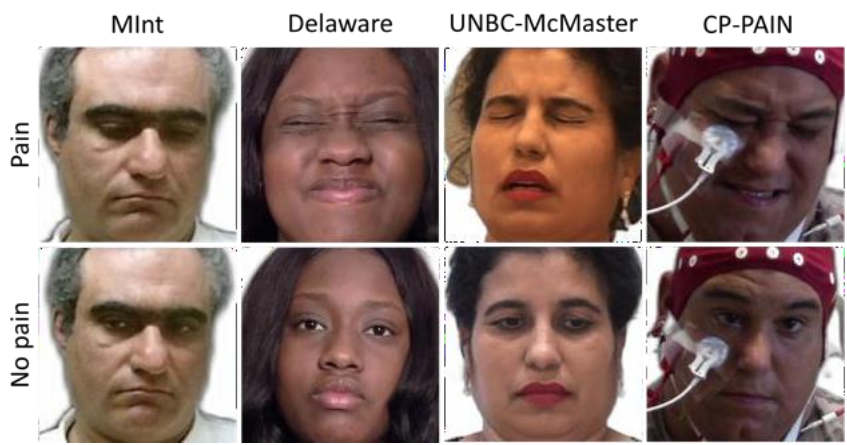

*Models*

We built ten neural networks aiming at recognizing pain from images (see Table 3). For the sake of simplicity, we denoted the models introduced in Song et al. (2014) 60., Li et al. (2015) 61, and Ramis et al. (2022) 43., as SongNet, WeiNet, and SilNet, respectively.

***Table 3: List of network architectures used in this study for pain recognition.***

| Work | Model/s |
| --- | --- |
| Krizhevsky et al. (2012) 62. | AlexNet |
| Song et al. (2014) 60. | SongNet |
| Li et al. (2015) 61. | WeiNet |
| Simonyan and Zisserman (2015) 63. | VGG16 & VGG19 |
| He et al. (2015) 64. | ResNet50 & ResNet101V2 |
| Szegedy et al. (2015) 65. | InceptionV3 |
| Chollet (2017) 66. | Xception |
| Ramis et al. (2022) 43. | SilNet |

All the employed networks share a foundation in convolutional architecture; however, they diverge in terms of architectural constituents, including the arrangement of convolutional layers, pooling layers, fully connected layers, and other elements. This architectural disparity is reflected in the count of parameters employed by each network.

Six out of the ten models - VGG16, VGG19, ResNet50, ResNet101V2, Xception, and InceptionV3 – were initialized with pre-trained weights sourced from the ImageNet dataset 67. The remaining models – AlexNet, SongNet, WeiNet, and SilNet – were trained from scratch.

By integrating these varied architectures, we can compare their performance and analyze the impact of architectural selections concerning tasks related to pain recognition.

The models were implemented using the Python programming language, harnessing the capabilities of the Keras library. In the case of the AlexNet, WeiNet, SongNet, and SilNet models, we constructed each layer sequentially within the Keras framework. Conversely, the remaining models were conveniently available through the Keras API, inclusive of their pre-trained weights sourced from the ImageNet dataset.

*Metrics*

To assess the performance of the different models, common metrics used when evaluating classification tasks, namely accuracy, precision, recall and F1-score, were used:

- Accuracy: it measures the proportion of correctly classified instances among the total instances in the dataset. It provides an overall view of the model's correctness. However, it might not be suitable when dealing with imbalanced datasets where one class dominates, as it can be misleading.

- Precision: it is a measure of how many of the instances predicted as positive by the model are actually true positives. It focuses on the correctness of positive predictions. High precision indicates that the model is careful in labeling instances as positive.

- Recall: it quantifies how many of the actual positive instances were correctly predicted by the model. It emphasizes the model's capability to capture all positive instances.

- F1-Score: it is the harmonic mean of precision and recall. It combines both precision and recall into a single value, providing a balanced assessment of a model's performance. It is particularly useful when dealing with imbalanced datasets or situations where both false positives and false negatives are of concern.

When evaluating a model's performance in image classification, these metrics collectively provide a comprehensive understanding of its strengths and weaknesses. Accuracy gives a global view of performance, precision focuses on correct positive predictions, recall emphasizes capturing all true positives, and F1-score offers a balanced perspective considering both precision and recall. Using this set of metrics ensures a nuanced assessment of a model's effectiveness in classifying images accurately and reliably, even in scenarios where data distribution might be uneven or where the cost of false positives and false negatives differs significantly.

*Models' explanation*

We applied the model-agnostic XAI technique known as Local Interpretable Model-agnostic Explanations (LIME) [68]. LIME operates by introducing perturbations to the image under examination, thereby generating multiple modified versions of the image. These perturbed samples are subsequently processed through the model, and the resulting prediction outcomes are employed to establish the relevance of each region within the image with respect to a specific class. To further understand the classification processes employed by the models, we generalized the local explanations derived from a subset of samples to formulate global explanations for each class. This approach followed the methodology introduced in Manresa-Yee et al. (2023) [69]., where local explanation masks are transformed into a normalized space and then averaged to produce a comprehensive heatmap that delineates the significance of individual facial features in the classification process. For a better comprehension of the explanation process, Figure 3 provides a visual representation of the distinct steps involved.

*Figure 3: Visualization of the explanation process followed for the identification of the important regions for a specific model to classify into a class.*

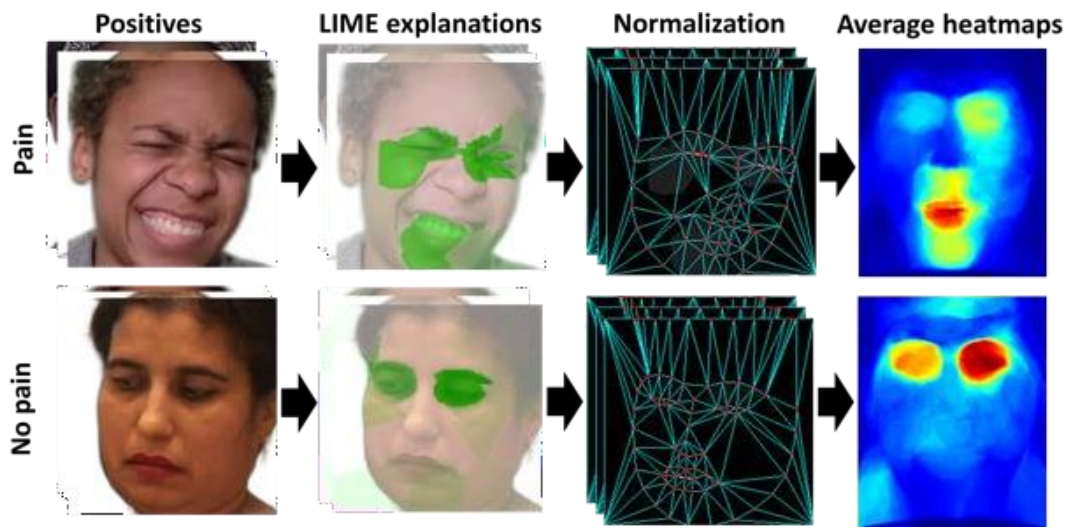

*Procedure*

All trainings for the ten networks were performed on a computer featuring an NVIDIA 4090 GPU and an i9 9900KF CPU, generously supplied by the Universitat de les Illes Balears. This same hardware configuration was consistently employed for the evaluation phase.

Following a series of preliminary experimental iterations, we empirically ascertained the hyper-parameter values for the various models. Notably, we set the number of epochs to 30 and the learning rate to 0.001. Additionally, data augmentation layers were incorporated into the training pipeline, encompassing randomized processes such as rotation, mirroring, and contrast adjustment. This integration aimed at introducing invariance to these intrinsic properties while concurrently augmenting the spectrum of image variations during training.

The experimental design comprised a total of three training scenarios for each of the ten models detailed in the Models Section. Further, we employed K-cross validation with K= 5 to ensure more stable and reliable results, for each of the three training scenarios, 150 trainings were performed in total.

Initially, the merged PAIN-DB dataset was partitioned into training and testing subsets, maintaining an 80%-20% and with varying test subsets, according to the K-cross validation procedure. This training phase involved models being trained on nearly all users' data, thereby assessing their performance in predicting pain based on images from users they had previously encountered. This training configuration aimed to yield the most optimal outcomes among the three.

The second training iteration closely resembled the initial one, using the PAIN-DB dataset. However, in this iteration, the partitioning was user-centric rather than image-centric, while preserving the same train-test ratio. Here, models were trained using images from a subset of users, with the remaining users reserved for the testing phase. This approach sought to evaluate the models' ability to accurately perceive pain within images of users unseen during training.

Finally, a training with the entire PAIN-DB dataset was conducted, testing on the pain images collected from individuals of the CP-PAIN dataset. The primary distinction here was that models were evaluated on users external to the PAIN-DB dataset. For this particular training scenario, there was no K-cross validation, but rather five different trainings on the same data, since the testing set is external to the PAIN-DB dataset.

The three-tiered approach to experimentation was devised to sequentially validate network performance on well-established pain recognition datasets. This stratification allowed for enhanced assessment of model effectiveness on a novel dataset, the size of which precluded fine-tuning strategies.

## RESULTS

### Inter-rater agreement among physiotherapists

In the course of the study, both physiotherapists conducted a grand total of 762 assessments, all stemming from the scrutiny of 127 video recordings. The inter-rater agreement, measured through the ICC, exhibited the following hierarchy of agreement levels for the observational measures:

- FACS = 0.751 (excellent)
- Wong-Baker Faces pain rating scale= 0.639 (fair/good)
- NCAPC = 0.551 (fair/good)

For a more comprehensive understanding of the inter-rater agreement contingent on the type of painful stimulus, refer to Table 4.

*Table 4: ICC of observational measures and pain/no pain measurement proportion. NCAPC: Non-Communicating Adults Pain Checklist; FACS: Facial Action Coding System; Wong-Baker: the Wong-Baker Faces pain rating scale*

| Source of pain | Measure | | |
|---|---|---|---|
| | FACS | Wong-Baker | NCAPC |
| Intramuscular injection | 0.412 | 0.590 | 0.565 |
| Muscular stretching | 0.845 | 0.739 | 0.891 |
| Other | 0.661 | 0.469 | 0.764 |
| Videos rated as "no pain" | 20 | 19 | 16 |
| Videos rated as "pain" | 107 | 108 | 111 |

These findings show a higher ICC across all scales when pain was induced by muscular stretching (mean of 3 scales = .825, excellent ICC), followed by an unknown source of pain (mean of 3 scales = .632, fair ICC), and lastly, intramuscular injection (mean of 3 scales = .522, fair ICC). Given that FACS exhibited the highest level of agreement, it was employed to label the image into the "pain"/"no pain" classes.

### Deep Learning pain recognition

The outcomes derived from the three distinct training scenarios are presented in Figure 4. As expected, the initial scenario, where networks were evaluated on users they had encountered during training, yielded the most favorable results overall. Notably, upon assessment with previously unencountered users, a marginal reduction was observed in both accuracy and F1 score across most networks. The third scenario, involving testing on users from the distinct CP-PAIN dataset, led to slightly diminished metrics. Among the models assessed, merely three attained performance levels surpassing 70%: InceptionV3, ResNet101V2, and Xception. Remarkably, InceptionV3 exhibited the most promising outcomes on the CP-PAIN dataset, attaining an accuracy of 62.67% alongside an F1 score of 61.12%.

*Figure 4: Accuracy (top) and F1 score (bottom) of the pain prediction results of the three different trainings performed, displayed by network.* The values correspond to the average between the five validation splits, for each training scenario.

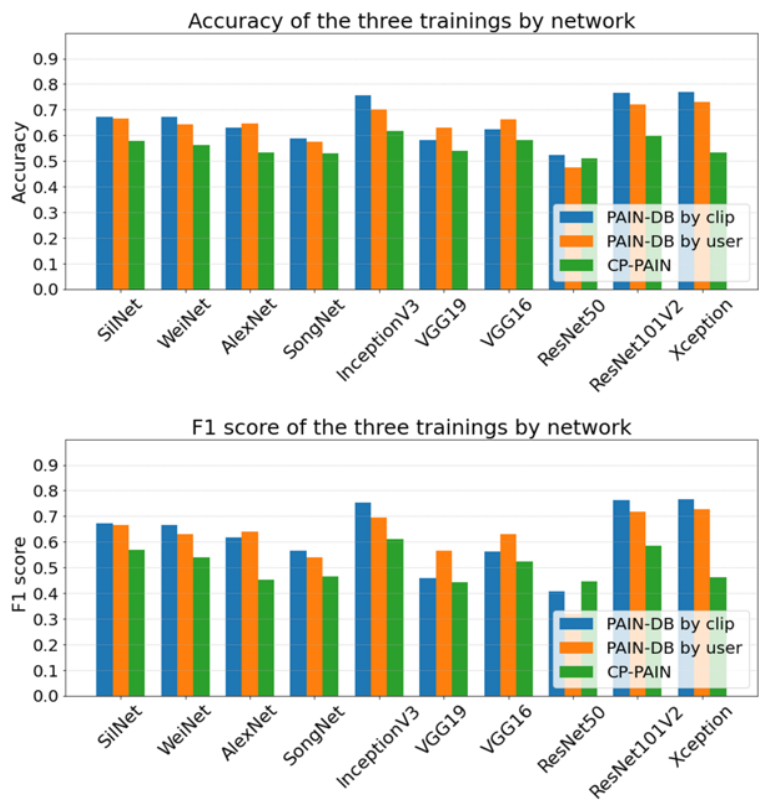

Figure 5 illustrates precision and recall values for each class (pain and no pain) on the CP-PAIN dataset. The precision values across networks display a relatively balanced distribution between classes. Interestingly, the recall values, which highlight the model's aptitude for identifying specific classes, underscore a consistent trend toward pain prediction over no pain prediction for all networks. This trend is particularly pronounced in certain models such as AlexNet, VGG19, ResNet50, and Xception, which achieve a recall of 80% or higher for pain but exhibit figures of 20% or lower for no pain instances. In contrast, InceptionV3, ResNet101V2, and SilNet exhibit more balanced recall values. This balance is further translated into higher F1 scores, as depicted in the lower chart of Figure 4.

**Figure 5: Precision (top) and recall (bottom) of the pain prediction results by class on the CP-PAIN dataset, displayed by network.** *The values correspond to the average between the five validation splits, for each training scenario.*

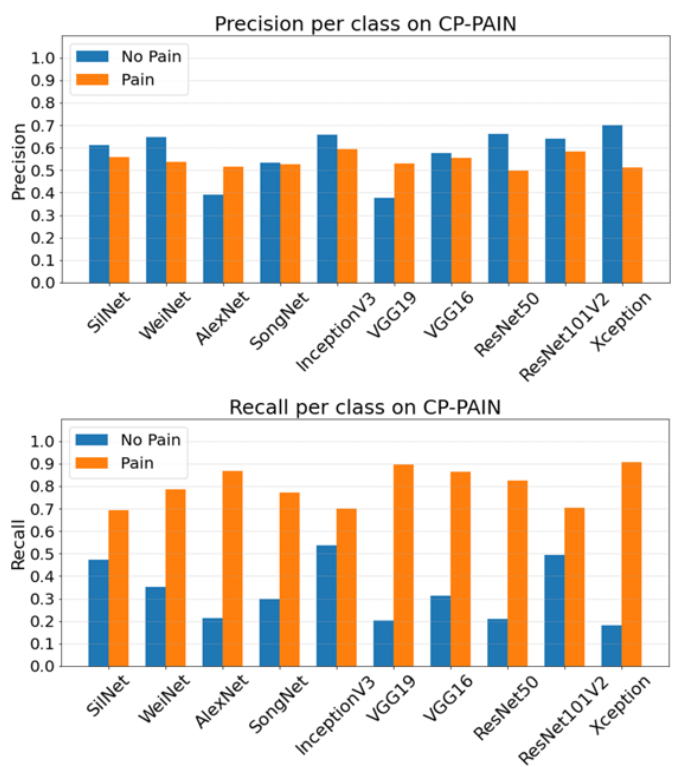

Figure 6 presents heatmaps that illustrate the significance of distinct facial regions in predicting a specific class for each of the trained models. Evidently, there exists subtle divergence among the models regarding the specific facial regions they emphasize when discerning the presence or absence of pain. However, noticeably, the heatmaps from models with pre-training and those without pre-training appear to segregate into two distinct clusters. This observation suggests a proclivity for models within each category to prioritize particular facial regions to a greater or lesser extent during the classification process.

*Figure 6: Heatmaps representing face regions importance for the different models, classes and databases.*

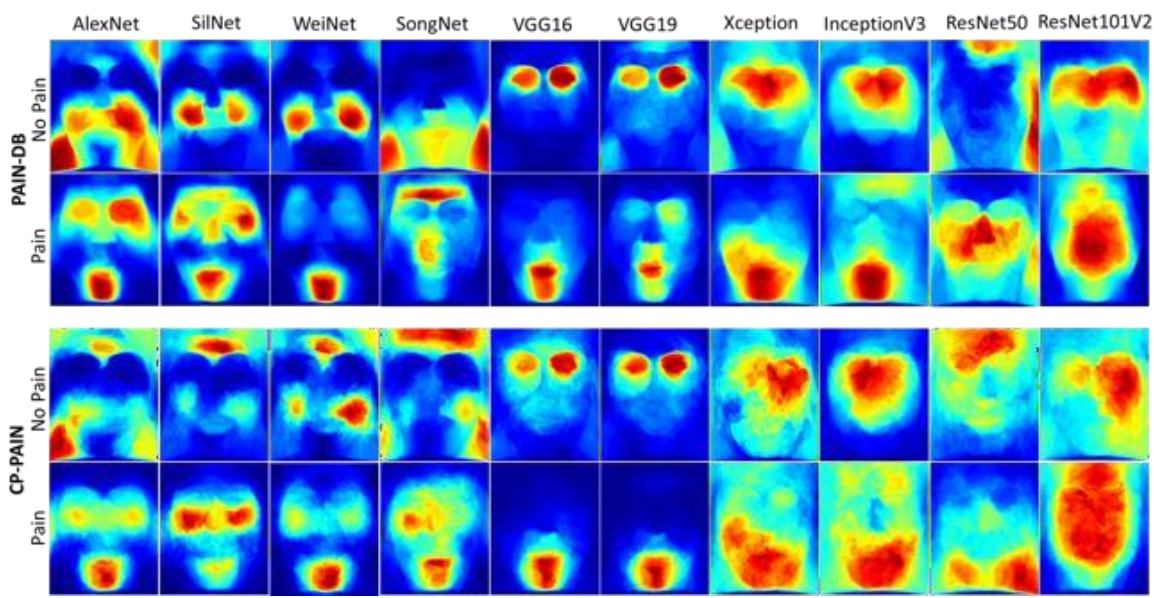

In an ideal scenario, the models should exhibit similar heatmaps when applied to both the PAIN-DB and CP-PAIN datasets. However, as depicted in Figure 6, certain models deviate from this ideal alignment. SilNet, WeiNet, and SongNet prominently emphasize the forehead region when classifying instances as "Not pain" in the CP-PAIN dataset, a pattern not evident in the PAIN-DB dataset. This disparity underscores the substantial impact of database differences on model predictions. Similarly, ResNet50 also demonstrates distinct results by focusing on the lower facial region for pain recognition in CP-PAIN, while primarily centering on the central facial area in the case of the PAIN-DB dataset. In contrast, the remaining models appear to exhibit relatively consistent reliance on the same facial regions, irrespective of the database under consideration. Another noteworthy observation pertains to the importance of the lower portion of the face, particularly encompassing the mouth and its vicinity, which emerges as a pivotal factor in pain recognition. Conversely, the absence of pain is predominantly associated with the upper facial region, particularly focusing on the eyes and their surroundings.

## DISCUSSION

The primary objective of this study was to develop an automated facial recognition system based on DL for the assessment of pain in adults with CP. To achieve this goal, we developed and trained this system using a specific dataset of images of individuals with CP (CP-PAIN) curated for this very purpose and three existing well-known pain databases. Subsequently, we compared pain scores obtained from the automated facial recognition system with the pain scores provided for each image of the CP-PAIN dataset, obtained by consensus by two independent physiotherapists experienced in caring for individuals with CP. Our findings revealed a 60% accuracy rate for the facial recognition system, thus confirming the feasibility of adapting pain detection from images for patients with CP.

To the best of our knowledge, this study marks a pioneering initiative introducing pain detection through image analysis for individuals with CP. This innovation has the potential to offer valuable solutions for evaluating pain within this population, a challenge often noted by clinicians and family members alike 12. Beyond addressing this pressing need, the application of image-based pain detection has the added advantage of mitigating the intrinsic subjectivity inherent in human pain assessments 20-22., thereby reducing associated inaccuracies. Our findings reflect a noteworthy aspect of this subjectivity. Even in cases where the evaluator has a deep familiarity with the individual with CP under assessment, our results revealed the presence of a subjective bias that depends on the evaluator's level of familiarity with the specific painful procedure being evaluated. To this regard, physiotherapists exhibited a higher level of agreement when assessing pain induced by muscle stretching compared to pain resulting from intramuscular injection or an unknown source of pain. This subjectivity in pain assessment has been demonstrated in other studies that measured the degree of agreement between individuals with cerebral palsy and their parents 24,25., as well as with healthcare professionals 26. This highlights the importance of objective, technology-assisted approaches in enhancing the precision and objectivity of pain assessment in individuals with CP.

Given the scarcity of relevant patient data for training, we devised a strategy involving the training of pain detection models on established datasets tailored for this task. Subsequently, the models' performance was evaluated on a minimal test set, CP-PAIN, constructed specifically for this study, which comprises images of users with CP, both with and without pain. Despite the limited number of images contained in CP-PAIN, this dataset is the first of its kind. Cognitive factors and motor impairment may affect the gestures, body movement and mimics in individuals with neuromotor disorders, such as individuals with CP, what could lead to idiosyncratic pain expressions or mask pain of low intensity 53,70. Our approach incorporates a specific dataset of pain in this population that may help enhance DL pain recognition. Further refinement and expansion of our dataset can harness the power of DL to create a more robust and reliable pain detection system. The need for a larger and more diverse image datset becomes evident as we strive to train our DL model to better recognize nuanced pain expressions in this unique population. Although limited and prone to expansion and quality improvement, the CP-PAIN database could be a first contribution to the DL analysis of pain expressions in complex populations with pain facial expressions diverse from those of the general population 53,70., collected in other existing databases.

Initially, the performance achieved on the training set reached a maximum of 70% for users unknown to the models. Consequently, this outcome can be considered an upper-bound estimation for the performance attainable on CP-PAIN. Acknowledging this modest performance, likely attributed to the limited training data, we plan to enhance these results in future work by diversifying the training datasets. Additionally, we emphasize the necessity for meticulous analysis of images from each database, particularly those extracted from video recordings, since the imprecision in annotations during video frame selection could introduce a significant number of erroneously labeled images, warranting careful consideration.

Our findings on the CP-PAIN dataset unveiled 60% accuracy, thereby establishing the viability of adapting pain detection from images to patients with CP. A noteworthy trend emerged where the models exhibited a pronounced trend towards detecting pain in images, surpassing instances of identifying no pain. This proclivity could be attributed to the frequent appearance of subjects in the test images exhibiting open mouths and exposed teeth, a phenomenon potentially associated by the networks with pain expressions seen during training. This accuracy in pain detection, while seemingly modest, marks a significant milestone in clinical research, as it underscores the potential for developing a specialized assessment model tailored to individuals with complex neuromotor conditions, a development that holds promise in reducing the inherent subjectivity and biases often associated with human evaluations of pain [20-22].

For the models exhibiting the highest performance, such as InceptionV3 and ResNet101V2, the application of XAI techniques have unveiled an interesting finding: when employing models trained on the PAIN-DB dataset on users with CP, the essential regions crucial for accurate pain identification remain largely unaffected. The consistent focus of these two models on identical facial regions for pain recognition both for users with CP and without suggests a noteworthy similarity in the expression of pain between these two distinct groups. However, it is essential to acknowledge that this finding does not consider potential idiosyncrasies in the facial expression of pain in subjects with complex neurological disorders, which may have led to the presence of false positives. It is important to note that some idiosyncratic behaviors, such as crying, moaning, flinching, having red cheeks, grunting, or sticking out the tongue can be misleadingly labeled as pain within the CP population, while they can express other physiological or emotional events [71]. On the contrary, other facial gestures, such as smiling or laughing, may be used to express pain in individuals with poorer communication or motor ability [72]. This underscores the need for further training of the system using a more diverse dataset of images of individuals with CP. We are committed to addressing potential variations in the expression of pain and enriching the robustness of our models for pain recognition across diverse user groups. This approach would ensure a more precise and reliable application of our technology in clinical and medical settings.

This study was not without limitations. In addition to the previously mentioned constraint of a limited number of images in our database, we also faced the unforeseen challenge of not being able to collect self-reports from most individuals with CP, as originally planned, what would have been valuable to evaluate the facial recognition system's measurements in a more accurate way. Another potential limitation was the unequal representation of different types of painful stimuli.

DL pain recognition allows envisioning the possibility of mitigating the limitations of human-based assessments of pain in complex conditions, such as CP, and achieving a level of objectivity and consistency that can significantly benefit the care and well-being of individuals with communication problems and cognitive and neuromotor disorders affecting pain expression. Therefore, our work not only showcases the potential of adapting technology for healthcare applications but also emphasizes the importance of ongoing research and data collection to advance and extrapolate the capabilities of our model in the future.

## Acknowledgments

The authors would like to thank the healthcare staff of the ASPACE Balearic Foundation and APACE Toledo for their various contributions during the development of this study.

This research was funded by MCIN/AEI/10.13039/501100011033 Spain, grants PID2020-114967GA-I00 (SENTS?) and PID2019-104829RA-I00,EXPLainable Artificial INtelligence systems for health and well-beING (EX-PLAINING) and for the Ministry of Science and Technology of Spain /Spanish Foundation for Science and Technology (FECYT,) grant FCT-20-16485).

In addition, we also acknowledge the funding of the FPU scholarship from the Ministry of European Funds, University and Culture of the Government of the Balearic Islands.

## Conflicts of interest

The authors declare that there is no conflict of interest.

## Abbreviations

AI: Artificial Intelligence.

CP-PAIN: dataset of facial pain expression images for individuals with cerebral palsy based on video recorded during potentially painful procedures.

CP: Cerebral Palsy.

FACS: Facial Action Coding System.

LIME: Local Interpretable Model-agnostic Explanations.

MInt PAIN: Multimodal Intensity Pain dataset.

NCAPC: Non-Communicating Adults Pain Checklist.

PAIN-DB: dataset including the UNBC-McMaster Shoulder Pain Expression Archive Database, the Multimodal Intensity Pain Dataset, and the Delaware Pain Database.